%% file: acl2023.tex
\pdfoutput=1

\documentclass[11pt]{article}

\usepackage[]{ACL2023}

\usepackage{times}
\usepackage{latexsym}

\usepackage[T1]{fontenc}

\usepackage[utf8]{inputenc}

\usepackage{microtype}

\usepackage{inconsolata}

\usepackage{amsmath}
\usepackage{amssymb}

\usepackage{tcolorbox}
\usepackage{xcolor}
\definecolor{darkpurple}{HTML}{6600CC}
\definecolor{darkgreen}{HTML}{009900}
\definecolor{darkyellow}{HTML}{CC9900}
\definecolor{lightpurple}{HTML}{A680B8}
\definecolor{back_gray}{HTML}{EDF3FF}

\usepackage{enumitem}

\usepackage{xspace}
\newcommand{\model}{\textit{LEFT}\xspace}
\newcommand{\modelwithspace}{\textit{LEFT}\xspace}

\newcommand{\upa}{\mathbf{A}}
\newcommand{\upb}{\mathbf{B}}
\newcommand{\upc}{\mathbf{C}}
\newcommand{\upu}{\mathbf{U}}
\newcommand{\upx}{\mathbf{X}}

\newcommand{\lowa}{\mathbf{a}}
\newcommand{\lowb}{\mathbf{b}}
\newcommand{\lowc}{\mathbf{c}}

\newcommand{\lowo}{\mathbf{o}}

\usepackage{lipsum}
\usepackage{booktabs}

\usepackage{algorithm}
\usepackage{algpseudocode}

\usepackage{multirow}

\usepackage{svg}
\title{Reverse That Number! Decoding Order Matters in Arithmetic Learning}

\author{Daniel Zhang-Li$^1$\thanks{\quad Equal contribution}, Nianyi Lin$^1$\footnotemark[1], Jifan Yu$^1$, Zheyuan Zhang$^1$, Zijun Yao$^1$, Xiaokang Zhang$^2$,\\
\textbf{Lei Hou$^1$, Jing Zhang$^2$, Juanzi Li$^1$\thanks{\quad Corresponding Author}} \\
  $^1$Department of Computer Science and Technology, Tsinghua University \\
  $^2$Renmin University of China \\
  \texttt{\{zlnn23,linny20\}@mails.tsinghua.edu.cn} \\
  \texttt{\{lijuanzi\}@tsinghua.edu.cn }
  }

\begin{document}
\maketitle
\input{data/abstract}

\input{data/Intro}
\input{data/problem}

\input{data/method}
\input{data/implementation}
\input{data/exp}

\input{data/ablation}

\input{data/related}

\input{data/conclusion}
\input{data/ethnic}

\bibliography{anthology,ref}
\bibliographystyle{acl_natbib}

\appendix


\end{document}

%% file: data/abstract.tex
\begin{abstract}

Recent advancements in pretraining have demonstrated that modern Large Language Models (LLMs) possess the capability to effectively learn arithmetic operations. However, despite acknowledging the significance of digit order in arithmetic computation, current methodologies predominantly rely on sequential, step-by-step approaches for teaching LLMs arithmetic, resulting in a conclusion where obtaining better performance involves fine-grained step-by-step. Diverging from this conventional path, our work introduces a novel strategy that not only reevaluates the digit order by prioritizing output from the least significant digit but also incorporates a step-by-step methodology to substantially reduce complexity. We have developed and applied this method in a comprehensive set of experiments. Compared to the previous state-of-the-art (SOTA) method, our findings reveal an overall improvement of $11.1\%$ in accuracy while requiring only a third of the tokens typically used during training. For the purpose of facilitating replication and further research, we have made our code and dataset publicly available at \url{https://anonymous.4open.science/r/RAIT-9FB7/}.

\end{abstract}

%% file: data/Intro.tex
\section{Introduction}

Large language models (LLMs), though proficient in a range of tasks~\cite{ouyang2022training,achiam2023gpt,team2023gemini}, encounter challenges in arithmetic operations due to their inherent design limitations, such as reliance on next-token prediction methods and limited working memory~\cite{bubeck_sparks_2023}. Despite their capability to utilize external tools for circumventing direct arithmetic computations during inference~\cite{gao2023pal,imani2023mathprompter,schick2023toolformer}, efficiently and effectively incorporating arithmetic proficiency within LLMs is an unresolved issue. However, previous studies have demonstrated that LLMs can learn arithmetic effectively through pretraining~\cite{yang_gpt_2023}. This suggests that it might be feasible to efficiently teach LLMs arithmetic operations through fine-tuning alone, without the need external tool such as calculators.


\begin{figure}[t]
  \centering
  \includegraphics[width=0.8\linewidth]{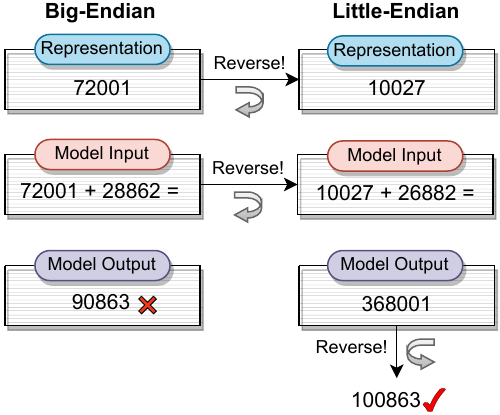}
  \caption{Reversing the numbers in training enables models to better learn to do arithmetic operations.}
  \label{fig:intro}
\end{figure}



The prevailing challenge in employing Large Language Models for arithmetic tasks is intricately linked to their next-token prediction mechanism. This mechanism often leads to a reversed computation order, where more significant digits are calculated before less significant ones, a flaw attributed to LLMs' inherent limitation in forward planning~\cite{bubeck_sparks_2023}. This characteristic has led to the perception that arithmetic in LLMs is akin to other complex symbolic and logical tasks, necessitating a similar approach~\cite{nye2021work}. Consequently, prior research has predominantly focused on the necessity of a step-by-step methodology, breaking down arithmetic into a series of sub-steps, as a critical strategy for addressing these challenges~\cite{wei_chain--thought_nodate, lee2023teaching}.

\begin{figure*}[ht]
    \centering
    \includegraphics[width=0.9\linewidth]{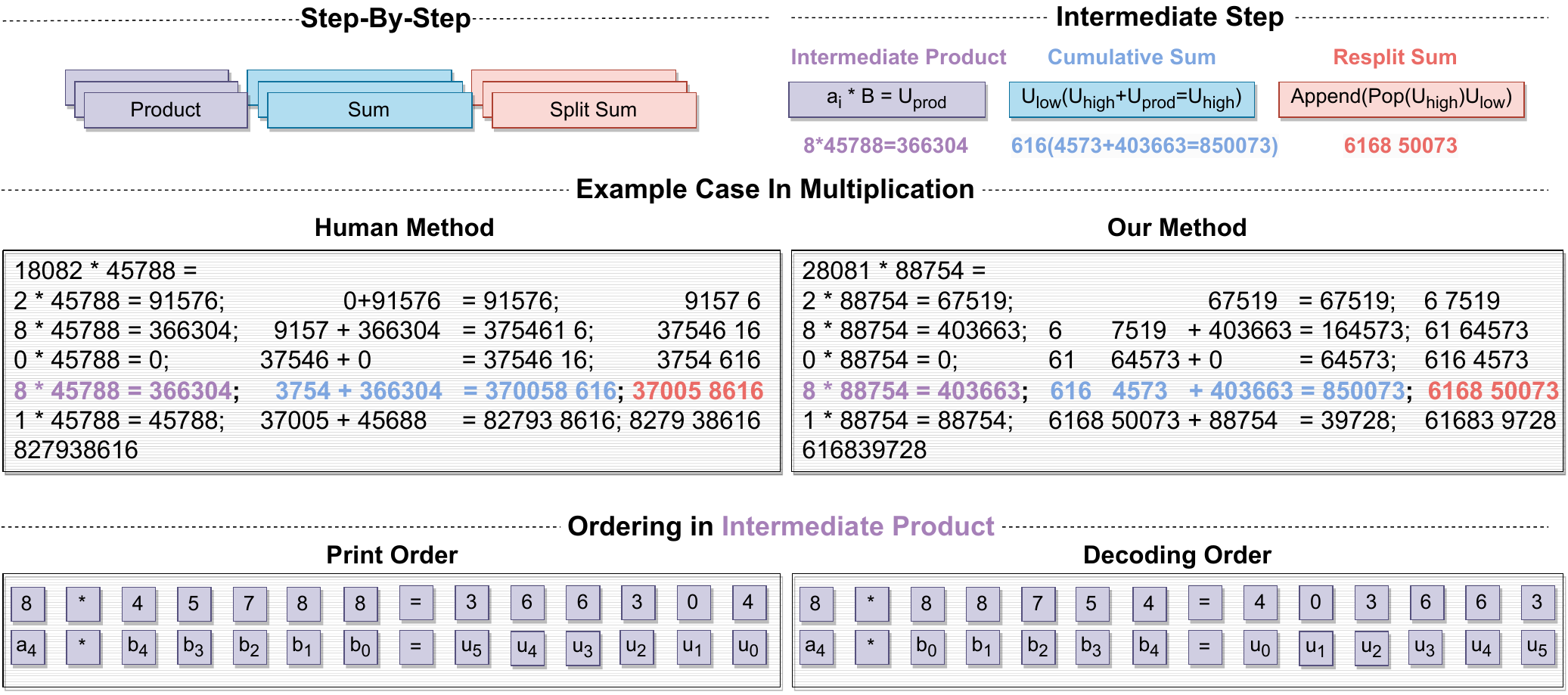}
    \caption{Example training data for Multiplication. Where the task is solved using a step-by-step process. During the $i$th intermediate step, the intermediate product is first computed. Then, inspired by the human process, we set the least significant digits($U_{high}$) unchanged and directly added the product to the remaining digits($U_{low}$) of the cumulative sum. Finally, we pop the least significant digit from the updated $U_{high}$ and append it into $U_{low}$ as it will not be added with non-zero digits in later steps. During decoding, we express all numbers in \textbf{Little-Endian}, where the least significant digit goes first. We convert all the numbers back to \textbf{Big-Endian} before printing.}
    \label{fig:example}
\end{figure*}

Such a technique achieves significant gains in performance but introduces a trade-off between efficiency and effectiveness, necessitating a balance between the number of tokens per training case and the total number of training cases. To enhance both efficiency and effectiveness without resorting to a brute-force integration of step-by-step processes, we adopt a novel approach termed \model(\textbf{L}ittle-\textbf{E}ndian \textbf{F}ine-\textbf{T}uning). Rather than incrementally integrating step-by-step mechanisms, we employ a strategy that reverses the number representation, prioritizing the computation of less significant digits. This approach utilizes the concept of \textbf{Little-Endian}, where numbers are represented with the least significant digits first, while maintaining the position of any negative signs. In contrast, the standard numeral representation is referred to as \textbf{Big-Endian}.
Figure~\ref{fig:intro} demonstrates that initiating output generation with the most significant digit may result in carry-related errors. In contrast, employing a Little-Endian format, where the model produces the number $100863$ as $368001$, simplifies carry operations resulting in a correct solution. We present experimental results (Sec.~\ref{sec:experiment}) showcasing that \modelwithspace not only improves accuracy by $11.1\%$ against the current state-of-the-art (SOTA) for large digit inputs but also demonstrates efficiency by utilizing just $5.2\%$ of the training tokens required by the previous SOTA for addition and subtraction tasks. Specifically, in multiplication, \modelwithspace records a $35.7\%$ performance gain while consuming only $56.6\%$ of the training tokens used by prior SOTA.
The key contributions of this paper include:
\begin{itemize}
    \item We proposed a novel method, \modelwithspace, leveraging Little-Endian to reduce the complexity of learning arithmetic operations.
    \item We conduct detailed evaluation and demonstrate \modelwithspace achieves better performance with lesser token used during training.
    \item Observations from our experiments indicate that, by reversing digit order, LLMs are capable of solving addition in human alike manner.
\end{itemize}

%% file: data/problem.tex
\section{Problem Formulation}

Consider the simple case where the input ($\mathcal{I}$) consists of two numbers, ${\upa}$ and ${\upb}$, combined with an operator ${op}$. We denote the digits of ${\upa}$ as ${{\upa}=\sum_{i=0}^{m-1}10^i{\lowa_i}}$, where each ${{\lowa_i}}$ is a single-digit integer ($0 \leq {{\lowa_i}} \leq 9$), and ${{\lowa_{m-1}}} \neq 0$ to ensure no leading zeros. Similarly, for ${\upb}$, we express its digits as ${{\upb}=\sum_{i=0}^{n-1}10^i{\lowb_i}}$, where each ${{\lowb_i}}$ is a single-digit integer ($0 \leq {{\lowb_i}} \leq 9$), and ${{\lowb_{n-1}}} \neq 0$.



We assume the ground truth output is a $k$-digit number, {$\upc = \sum_{i=0}^{k-1}10^i\lowc_i$} (for $\upc < 0$, we use{$\lowc_{-1}$} to represent the negative sign). The trained LLM outputs an ordered sequence $\mathcal{O}=\{\lowo_{1},\lowo_{2},\ldots\}$, which includes the output number ${\upc} \subseteq \mathcal{O}$.

As step-by-step designs often incorporate intermediate results, we denote the $i$th intermediate result as {$\upu^i$}. 
Finally, we define the remaining output as auxiliary tokens ($\upx = \mathcal{O} \setminus \{{\upu^i} \mid \forall i\} \cup \{{\upc}\}$).

%% file: data/method.tex
\section{Little-Endian Fine-Tuning}
\label{sec:method}


In order to effectively and efficiently teach LLMs arithmetic, we need to address three crucial questions: 1. What is the complexity in standard Big-Endian training(where no step-by-step is applied)? 2. Are there spaces for optimizing the standard method? 3. How to optimize cases when step-by-step is required? In the remaining parts of this section, we tackle such questions one by one.

\subsection{Learning Complexity of Arithmetic}




Autoregressive LLMs are interpreted as probabilistic models that predict output sequences by maximizing the likelihood of generating the correct output. In operations such as addition, this process of prediction can be formalized as follows:
\begin{equation}
\arg\max_{c_i}P(c_i|a_{0\sim n-1},b_{0\sim m-1},c_{i+1\sim k}) \nonumber
\end{equation}

Considering the specific nature of addition, where the outcome of each digit is influenced only by digits of equal or lesser significance, the process is refined to concentrate on pertinent inputs:

\begin{equation}
\arg\max_{c_i}P(c_i|a_{0\sim i},b_{0\sim i})
\end{equation}

Assuming that all numbers involved possess an identical number of digits simplifies the analysis. Under this assumption, during the generation of each digit, there exist $10$ potential inputs from each of the two numbers, resulting in $10^{2i+2}$ possible input combinations. Given that the output digit can assume 10 possible values, the complexity of predicting a single digit's value transitions from \(10^{2i+2}\) input conditions to 10 output conditions.

The overall learning complexity is quantified by summing the probabilities of accurately predicting each digit, based on the inputs up to that digit:

\begin{equation}
\mathcal{L}_{Big} = -\sum_{i=0}^n \log P(c_i|a_{0\sim i},b_{0\sim i})
\end{equation}

Accordingly, the cumulative learning complexity, denoted as \(\mathcal{C}_{Big}\), is conceptualized as the aggregate of complexities across all digits, with the input variations providing a lower bound:

\begin{equation}
\mathcal{C}_{Big} = \sum_{i=0}^n 10^{2i+2} \geq 10^{2n+2}
\end{equation}

This model illustrates the exponential increase in learning complexity with the increment of digit count \(n\), presenting a significant scalability challenge in teaching arithmetic to LLMs.

\subsection{Optimizing Complexity via Little-Endian}
\label{sec:left-e2e}
In addressing the complexity of arithmetic operations, it is noted that the output token with the greatest complexity is typically the most significant digit. Interestingly, unlike computational models, humans often do not consider all input digits simultaneously. Instead, they start from the least significant digit, using any carry-over to simplify the computation. Assuming the model can similarly infer the carry from the previous digit ($a_{i-1}, b_{i-1}, c_{i-1}$), we can streamline the optimization target by focusing on this simplified context:
\[
\arg\max_{c_i} P(c_i|a_i, a_{i-1}, b_i, b_{i-1}, c_{i-1})
\]

Such adjustment leads to a significant reduction in input complexity, now quantified as $10^{5}$. By adopting this revised generating order, the task becomes markedly less challenging:
\[
\mathcal{C}_{Little}=\sum_{i=0}^n 10^{5} \leq n \cdot 10^{5}
\]

For cases where $n \geq 2$, this model showcases a substantial decrease in learning complexity compared to the conventional approach ($\mathcal{C}_{Little} \leq n \cdot 10^5 < 10^{2n+2} \leq \mathcal{C}_{Big}$). Such findings illuminate the potential benefits of inverting the decoding order to mitigate complexity. Motivated by this insight, we propose abandoning the classic, step-by-step design prevalent in previous methodologies in favor of revising addition and subtraction training to leverage this more efficient strategy.




\paragraph{Addition.} In addressing addition within \model, the traditional approach of processing numbers from the most significant digit to the least significant is reimagined. By reversing both the input and output numbers, the calculation aligns with the Little-Endian format, where operations commence from the least significant digit and progress towards the most significant. Such conversion simplifies the decoding order, making it more intuitive and akin to human arithmetic practices. We hypothesized that the model can autonomously recompute the necessary carry for the subsequent significant digit. This method eliminates the need for a step-by-step design or the introduction of auxiliary tokens, streamlining the addition process without necessitating any extra tokens beyond the sum itself.


\paragraph{Subtraction.} For subtraction, the model simplifies the process by first determining if the result will be negative, then applying the operation in Little-Endian order. This approach, which keeps the negative sign's position unchanged (e.g., -256 becomes -652), enhances efficiency by eliminating the need for intermediate results that assume a non-negative outcome. This streamlined method contrasts with traditional digit-wise subtraction, offering a more straightforward computation strategy.

\subsection{Augmenting Step-by-Step}
\label{sec:left-cot}

The application of Little-Endian formatting extends beyond the realms of addition and subtraction, offering substantial benefits in operations that inherently require a step-by-step approach due to their complexity. One prime example of such an operation is multiplication, where the intricacies of the computation process are significantly amplified.

\paragraph{Multiplication.}
Traditional methods often involve breaking down the solving process into manageable chunks, typically computing the product of a single digit with a multi-digit number, and then summing these intermediate products. This conventional approach, however, often operates under the Big-Endian framework, starting with the most significant digits and potentially complicating the computation of intermediate products.

In contrast, the use of Little-Endian proposes a significant optimization. By reversing the order of digits—starting from the least significant—this method aligns with the natural flow of human computation, simplifying both the computation of intermediate product and subsequent sums.

%% file: data/implementation.tex
\section{Implementation}
In this section, we delve into the detailed implementation of \modelwithspace and explore the methodologies applied in our experiments, along with the baselines for comparison. Our discussion spans from the step-by-step design utilized in the experiments (Sec.~\ref{sec:imp-cot}) to dataset generation (Sec.~\ref{sec:exp-dataset}) and other settings for the experiments(Sec.~\ref{sec:exp-setting}).

\subsection{Step-By-Step Design}
\label{sec:imp-cot}

\paragraph{Addition/Subtraction.} While our hypothesis posits that the step-by-step process might not be essential for efficiently learning addition and subtraction, we incorporate it as a comparative measure to validate our assumption. We adopt the step-by-step design from the chain-of-thought methodology~\cite{wei_chain--thought_nodate}, as reproduced in previous studies~\cite{zhou_teaching_2022}, for \modelwithspace's addition and subtraction tasks when necessary for evaluation.

\paragraph{Addition/Subtraction.} Contrary to our initial hypothesis that a step-by-step process may not be crucial for efficiently mastering addition and subtraction, we included it for comparative analysis to test our theory. Thus, we utilized the \textit{Chain-Of-Thought} approach~\cite{wei_chain--thought_nodate}, as previously replicated~\cite{zhou_teaching_2022}, in evaluating \modelwithspace joined with step-by-step on addition/subtraction.

\paragraph{Multiplication.} We previously outlined the key features of the step-by-step approach for multiplication within \modelwithspace, yet a direct implementation was not provided. As shown in Figure~\ref{fig:example}, with the reversal of all numbers, the task is divided into numerous substeps. Each substep iterates over the digits of the first input number, \(\lowa_i\in\upa\), starting from the least significant digit. In each iteration, the process begins by multiplying the current digit with the second input number to generate an intermediate product. This intermediate product is then added to the cumulative sum of products from previous iterations. Since the lower \(i\) digits of the product are always zero, these are not explicitly represented; instead, the product is directly added to the higher section of the cumulative sum. The higher section is defined as the part of the cumulative sum obtained in the last step of the previous iteration, which considers the lower \(i\)-digits as a fixed result and defines the remaining digits as the higher section of the cumulative sum.

This refined step-by-step design for multiplication highlights the efficiency and adaptability of the Little-Endian approach in managing complex arithmetic operations. By streamlining the integration of intermediate products into a simplified cumulative sum, this method not only improves the performance and clarity of the model but also showcases the extensive utility of Little-Endian formatting in enhancing computational processes.

\subsection{Dataset}
\label{sec:exp-dataset}
The inherent characteristics of arithmetic calculations, which do not necessitate human-generated labels, enable the automated generation of training and testing sets in our study. Our primary objective is to create a dataset that is fair, isolated, and balanced, facilitating a comprehensive evaluation of the \modelwithspace's effectiveness and efficiency.

\paragraph{Fairness.} Given that different methods may operate on varied data inputs, we aim to minimize the variance in performance attributable to different inputs as much as possible. To achieve this, we initiate the process by generating a set of \textit{meta} data during the data generation phase. Each piece of meta data is conceptualized as a triplet in the form \textit{$(\upa, op, \upb)$}. This triplet serves as a unified seed for generating training and testing data for each method, ensuring that the same set of input is utilized across methods. Then, each triplet is expanded and formatted to suit the specific requirements of each method's data format.

\paragraph{Isolation.} Recognizing the critical importance of preventing data leakage, we take meticulous steps to ensure the uniqueness of input number sets, denoted by $\{\upa, \upb\}$. This strategy guarantees that the test set contains no identical input number pairs as found in the training set, thereby also ensuring the uniqueness of each training and testing set.

\paragraph{Digit Distribution Balancing.} Echoing previous methods that have highlighted the importance of balanced data distribution~\cite{lee2023teaching}, we ensure that both the training and test sets are balanced such that the maximum quantity of any single number in each data slice falls within the digit range of $[5,12]$. Specifically, we generate in total of $15K$ training data and $3K$ test data, with $5K$ points for each operation, accompanied by $1K$ test data points for each operation, to maintain this balance.


\subsection{Experiment Setup}
\label{sec:exp-setting}
\paragraph{Baseline.}
We first include \textit{End-To-End} training used in during pretraining methods~\cite{yang_gpt_2023} as a ground to compare performance in previous methods.
We then include \textit{Scratchpad}\cite{nye2021work}, one of the early founders in using step-by-step approaches to break down arithmetic into multiple steps. We also include \textit{Chain-Of-Thought}~\cite{wei_chain--thought_nodate} which provided a general approach of breaking step-by-step to a wide range of complex tasks. In addition, we include the \textit{Detailed-Scratchpad} method introduced in \cite{zhou_teaching_2022}.
\cite{zhou_teaching_2022} also introduces \textit{Algorithmic-Prompting} technique but as it requires too many auxiliary tokens making it hard to fit 12-digit training into the context length. As a result, we exclude it during our evaluation.

\paragraph{Metric.}
As arithmetic reasoning is strongly affected by error propagation, solutions with intermediate errors are almost impossible to provide the correct solution. As a result, we directly use the accuracy (ACC) of the predicted output to evaluate the effectiveness of the methods. As the discussion for efficiency is aimed at training better-performed models using fewer resources, we record the amount of tokens used for training and observe the change in accuracy as more tokens are used.


\paragraph{Backbone Model.}
The base checkpoint for our experimental framework is Llama2-13B~\cite{touvron_llama_2023}, chosen for its status as a well-regarded and openly accessible LLM. To address the need for processing longer sequences, the model's context length has been extended to $4,096$ tokens.

%% file: data/exp.tex
\section{Experiments}
\label{sec:experiment}
We now turn to a systematic evaluation of the proposed method.
Specifically, we design and conduct a series of comprehensive analysis which seeks to answer the following research questions:
\begin{enumerate}[label=\textbf{Q\arabic*}, itemsep=0pt, parsep=0pt]
  \item \textit{Is \modelwithspace effective and efficient?}(Sec.~\ref{sec:exp-overall})
  \item \textit{What grants \modelwithspace the ability to effectively tackles the provided task?}(Sec.~\ref{sec:exp-case})
  \item \textit{What can be further done on \modelwithspace?}(Sec.~\ref{sec:exp-digit})
\end{enumerate}


\begin{table*}[ht]
    \centering
    \small
    \begin{tabular}{l|cc|ccc|cr}
        \toprule
        \textbf{Method} & \textbf{Endian} & \textbf{StepByStep} & $+$ & $-$ & $\times$ & \textbf{Overall} & \textbf{Token Usage}\\
        \midrule
        \textit{End-To-End} & Big & No & 63.3 & 32.3 & 00.0 & 31.9 & 494,815\\
        \textit{Chain-of-Thought} & Big & Yes & 88.0 & 83.5 & 08.2 & 59.9 & 4,938,148\\
        \textit{Scratchpad} & Big & Yes & 94.8 & 73.1 & 00.0 & 56.0 & 5,747,670\\
        \textit{Scratchpad-Detailed} & Big & Yes & \textbf{99.8} & \textbf{97.3} & \underline{52.8} & \underline{83.3} & 10,995,191\\
        \midrule
        \model(Our) & Little & Mix & \underline{98.8} & \underline{95.9} & \textbf{88.5} & \textbf{94.4} & 3,040,616\\
        \bottomrule
    \end{tabular}
    \caption{Performance comparison between methods, trained with 5K data for each operation with randomly generated data. The maximum digits of input numbers for each data are equally distributed in the range of $[5,12]$ for each operation. The test set is generated in a similar manner but with only 1K data per operation. \modelwithspace uses Little-Endian to represent all numbers and excludes the step-by-step process for addition and subtraction.}
    \label{tab:main}
\end{table*}

\subsection{Direct Evaluation Over Performance}
\label{sec:exp-overall}
We began our analysis with the overall performance of \modelwithspace against previous methods for jointly trained and evaluated addition, subtraction, and multiplication performance. We then conduct operation-by-operation analysis to observe the results of training when jointly training is opt-out.

\paragraph{Observation 1: \modelwithspace Learns Faster Than Baselines.}
Table~\ref{tab:main} shows the resulting performance of each method after training. We order the baselines according to token used during training.
\modelwithspace used the least amount of training token among all the step-by-step methods, yet achieving $11.1\%$ performance improvement over previous SOTA.


Specifically, \modelwithspace's accuracy on addition and subtraction is slightly below \textit{Scratchpad-Detailed}. However, \modelwithspace only used $160K$ and $161K$ tokens for learning addition and subtraction. But \textit{Scratchpad-Detailed} used $2,936K$ and $3,254K$ for training. This means \modelwithspace uses only $1/20$ of training data yet still achieves similar performance. \modelwithspace also achieved $35.7\%$ accuracy improvement over previous SOTA on multiplication, further highlighting \modelwithspace's effectiveness and efficiency.

%% file: data/ablation.tex
\paragraph{Observation 2: Using Little-Endian Alone Obtains Better Efficiency On Addition/Subtraction.}
During method design(Sec.~\ref{sec:left-e2e}), we proposed that Little-Endian is a better substitute than existing methods, which leverage step-by-step to reduce the complexity required for arithmetic. However, we have not yet examined such a statement. This raised two major questions: (1) Would it be better to contain step-by-step? (2) How does step-by-step itself perform? As a result, we apply step-by-step for closer observation. We scale down the training data to half and a quarter of training cases than the joint evaluation and observe the change in performance. To omit influences caused by joint training, we train addition and subtraction separately. 

\begin{figure}
    \centering
    \includegraphics[width=0.49\linewidth]{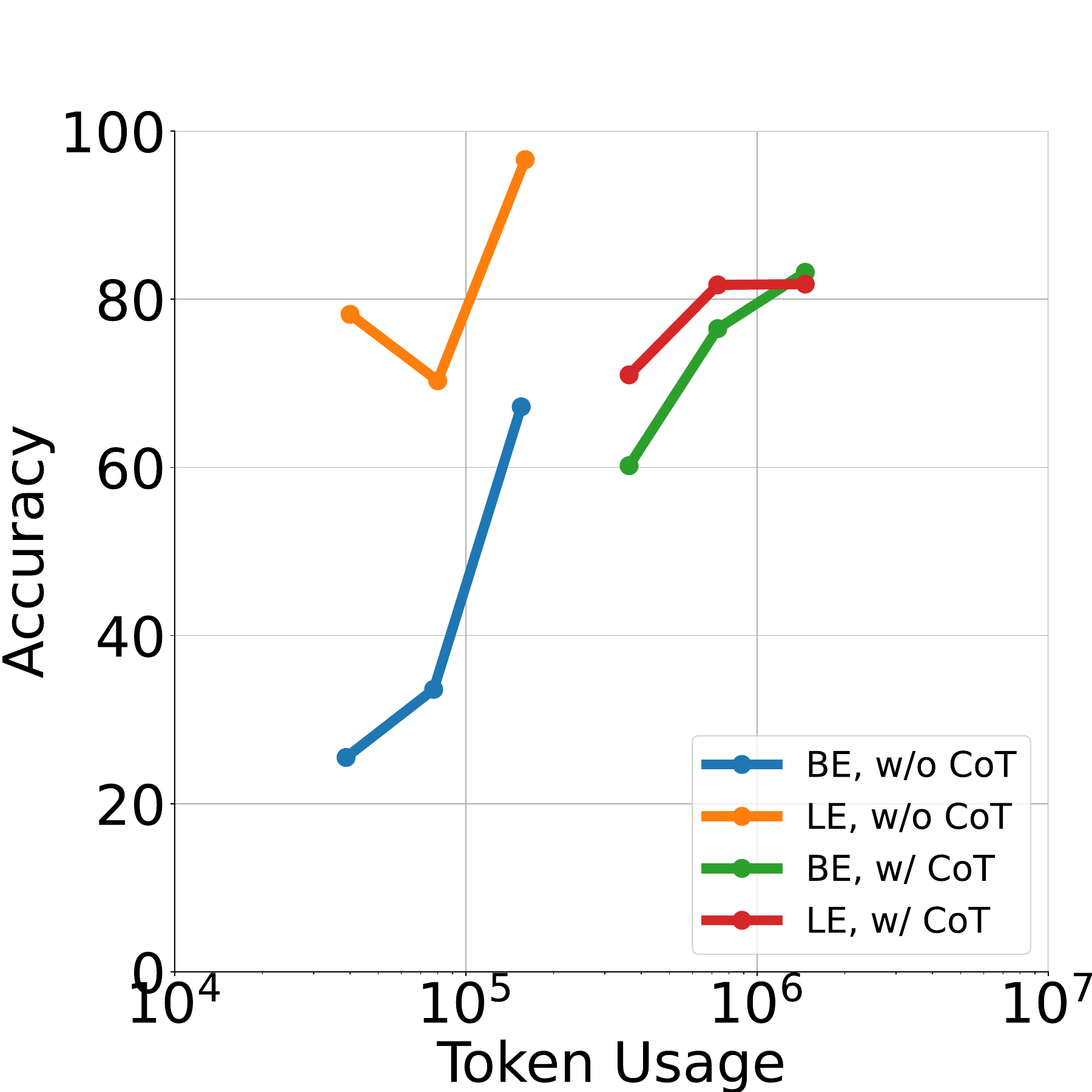}
    \includegraphics[width=0.49\linewidth]{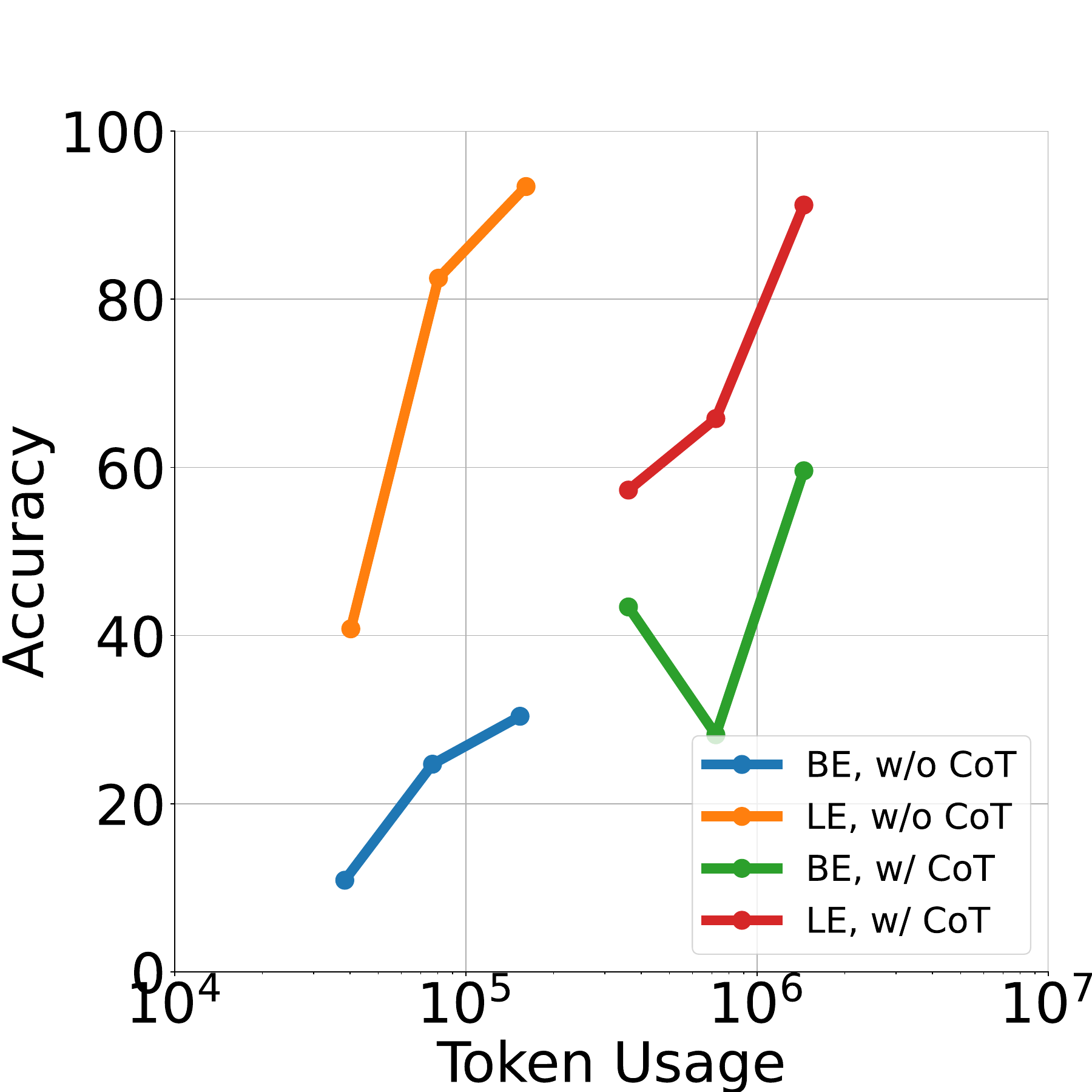}
    \caption{Performance when integrating step-by-step. BE stands for Big-Endian and LE stands for Little-Endian. The graph on the left shows the results after training on addition. The the right figure shows results for trained and evaluated on subtraction.}
    \label{fig:ablation-addition}
\end{figure}

As shown in Figure~\ref{fig:ablation-addition}, we observe that the use of Little-Endian outperforms other settings in both operations, despite the use of fewer tokens when compared to the step-by-step settings.


Moreover, we observe that the conventional \textit{Chain-Of-Thought} approach, which does not incorporate Little-Endian formatting, also significantly lags behind the \modelwithspace configuration. This outcome suggests that employing a step-by-step methodology does not invariably enhance performance. Particularly in addition, both the presence and absence of Little-Endian in the settings lead to inferior results compared to employing Little-Endian without a step-by-step approach. This implies that reversing the endian inherently captures critical information, which the step-by-step process aimed to convey in digit generation. Consequently, not only does the step-by-step application decrease efficiency, but it also deteriorates model performance by introducing additional chance of error propagation.

On the other hand, by taking a closer observation of subtraction, we see whether the use of step-by-step is integrated or not, the integration of Little-Endian brings much better performance. However, the learning curve of Little-Endian without step-by-step is smoother than in addition. We believe this could be related to the pretraining setting, where the model is trained with \textbf{Big-Endian}. On addition, when the carry is not occurring, knowing what endian is involved doesn't have a strong effect on the result, the model could falsely interpret the task as aligning the numbers with the leftmost digit and still achieve some level of performance. However, on subtraction, the endian greatly affects the result, as whether the result is negative is affected by the most significant digit, which is strongly related to the endian. Such difference resulted in poor performance in the beginning, as the model will have a great chance of failing unless it actually understands the task. But it also brings faster learning as the chance for the model to falsely understand the task reduces.
We believe such case highlights that the arithmetic ability of a fine-tuned model could be further improved with a backbone model that is pretrained with Little-Endian representation.


\begin{table}[h]
\small
\centering
\begin{tabular}{l|ccc|r}
\toprule
\multirow{2}{*}{\textbf{Method}} & \multicolumn{3}{c|}{\textbf{\# of Epochs}} & \textbf{Token}\\
& \textbf{1} & \textbf{2} & \textbf{3} & \textbf{Usage} \\
\midrule
\textit{End-To-End} & - & - & - & 186K \\
\midrule
\textit{Detailed-Scratchpad} & 24.9 & 32.6 & 39.3 & 4,805K\\
\midrule
\modelwithspace & 61.1 & 89.1 & 91.6 & 2,719K \\
\quad \textit{w/o Step-by-Step} & - & - & - & 186K \\
\quad \textit{w/ Big-Endian} & 24.2 & 42.8 & 52.7 & 2,719K \\
\bottomrule
\end{tabular}
\caption{Multiplication scores by different epochs and token usage. We observe settings without step-by-step solution failed to learn the task.}
\label{tab:ablation-multiplication}
\end{table}

\paragraph{Observation 3: Little-Endian And Step-by-Step Are Both Crucial For Multiplication.}

\noindent We now conduct a detailed examination for multiplication. We re-evaluate our backbone model to examine our designs on multuplication. For better comparison, we include two additional settings other than the standard \textit{End-To-End}. We first include a similar design as we proposed for solving addition and subtraction, where the model directly outputs the result but the input and output are both in Little-Endian. We then include \modelwithspace's step-by-step design but convert the numbers into Big-Endian. We also measure the different performances after different epochs of training to observe the convergence for the same amount of training cases.

The results are shown in Table~\ref{tab:ablation-multiplication}. We first observe that when the use of step-by-step is removed, it becomes impossible to learn multiplication. This demonstrates the need for step-by-step to break down the complexity in solving multiplication is still needed when only $5K$ of training data is available. We also observe that when Little-Endian is removed, the performance further improves over the step-by-step setting. The model also converges much faster, as the performance after $2$ epochs of training is already close to the performance of the last epoch, an accuracy of $91.6\%$. We are amazed that \modelwithspace achieves better performance when the model is trained only on multiplication, suggesting the potential for further optimization.

We also observe the number of tokens used during {\model}'s training in multiplication is approximately half of the tokens used by \textit{Scratchpad-Detailed}. In addition and subtraction training, tokens are better off with a factor of $20$. This shows that \modelwithspace with better performance achieves even greater improvement in token efficiency.


\subsection{Case Studies}
\label{sec:exp-case}
We now conduct a detailed study of the results obtained in the previous section, seeking to discover findings that can help future studies.

\paragraph{Finding 1: Little-Endian Reduces Step-By-Step Errors.}
In this section, we conduct an error analysis for the errors in our main experiment in order to find an explanation of the performance gain caused by changing the endian. To do so, we first selected the place where the first error occurred as an indication of the error of each falsely inferred test case. This is because error propagation is critical in arithmetic. We then focused on two crucial parts during each inference step, calculating the intermediate 1-by-$n$ product and the cumulative sum. As a result, we find that among the $417$ errors that occurred during intermediate calculations in \textit{Scratchpad-Detailed}: 1. $140$ errors occurred during calculating the intermediate product; 2. $236$ errors occurred during accumulating sum. Both operations had much better performance in \modelwithspace, where only $77$ errors were observed during computing the intermediate product and only $22$ errors were observed when updating the cumulative sum. The error occurrence is decreased by a factor of $10$ for summation and by a factor of $2$ for the intermediate product. We believe this is because the carry is easier than to compute when the less significant digits are already shown, which possibly could reduce the complexity in computing the result for the current digit. The error for the intermediate sum is reduced by a greater factor as the addition training is transferable when accumulating sum on \modelwithspace, whereas in \textit{Scratchpad-Detailed}, the addition task stands more on its own. Despite slightly better performing while evaluated on addition, it cannot transfer its ability to other tasks like multiplication.


\begin{figure*}[htbp]
    \centering
    \includegraphics[width=0.4\linewidth]{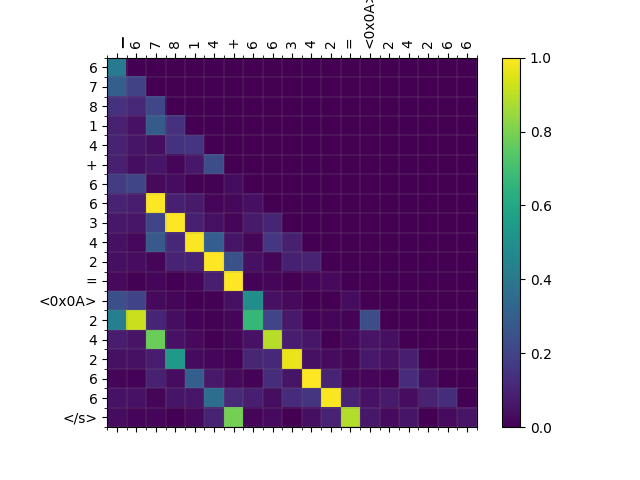}
    \includegraphics[width=0.4\linewidth]{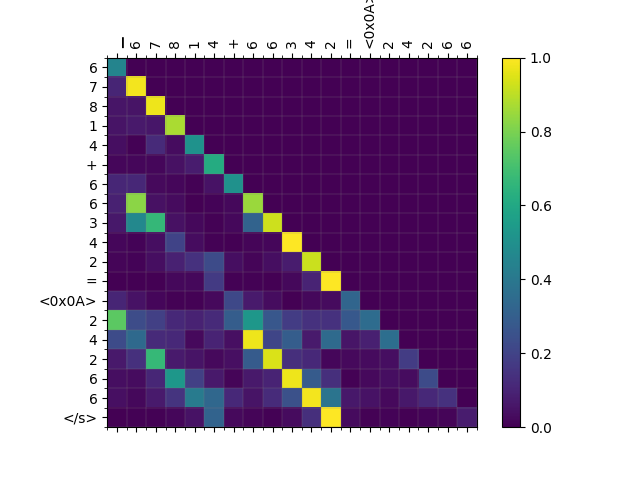}
    \caption{Visualization of attention weights during inference, with rows representing output tokens and columns indicating input tokens involved in generation. Attention weights are square-root transformed for enhanced visibility of correlations. The attention on the left(layer $14$) reveals output digits are correlate with their inputs, while attention(right) from layer $22$ suggests carry information reconstruction.}
    \label{fig:prob}
\end{figure*}

\paragraph{Finding 2: \modelwithspace Conducts Addition Just Like Humans}
We now take a closer observation of how \modelwithspace conducts addition. By logging the attention~\cite{vaswani2017attention} scores in the model, we observe a correlation between the output digit and related digits from the input numbers, as shown in Figure~\ref{fig:prob}. We observe that the input digits are recognized when computing the corresponding output during generation in some attention heads. We also observed that, in the $22$th layer, shown traits suggest the fine-tuned LLM has learned to re-compute the carry from the previous digits. Adressing our hypothesized during the method design, this proofs the assumption that the model can recover the carry when it's used (Sec.~\ref{sec:left-e2e}). This is a interesting indication because it suggests Little-Endian might be conducting training in a manner similar to how humans conduct addition without a draft paper.

\begin{table}[h]
    \centering
    \setlength{\tabcolsep}{2.2pt}
    \scalebox{0.83}{
    \begin{tabular}{lrrrrrrrrrrr}
    \toprule
    \textbf{Max Digit} 
    & \multicolumn{1}{c}{\textbf{5}} 
    & \multicolumn{1}{c}{\textbf{6}}
    & \multicolumn{1}{c}{\textbf{7}}
    & \multicolumn{1}{c}{\textbf{8}}
    & \multicolumn{1}{c}{\textbf{9}}
    & \multicolumn{1}{c}{\textbf{10}}
    & \multicolumn{1}{c}{\textbf{11}}
    & \multicolumn{1}{c}{\textbf{12}} \\ \midrule
    $+$                  & 100.0           & 98.4           & 100.0           & 99.2           & 97.6           & 97.6            & 98.4            & 99.2            \\
    $-$               & 92.0           & 96.8           & 93.6           & 96.8           & 100.0           & 100.0            & 93.6            & 94.4            \\
    $\times$            & 93.6           & 96.0           & 86.4           & 96.0           & 88.0           & 86.4            & 84.8            & 76.8             \\ \bottomrule
    \end{tabular}
    }
    \caption{Accuracy trends with increasing max input digits. We observe a steeper decline in multiplication's performance compared to other operations.}
    \label{tab:ablation-digit}
\end{table}

\subsection{Additional Error Analysis}
\label{sec:exp-digit}
Finally, we look at the errors occurred in \modelwithspace's joint experiment in the perspective of different maximum amount of input digits. As shown in Table~\ref{tab:ablation-digit}, \modelwithspace is able to perform well in lower digits, but when it is challenged towards higher digits of inputs, it loses part of its performance. Such a drop in performance is mostly significant when it comes to higher-digit multiplications, the digits being operated become much more complicated comparing to addition and subtraction. This stated that, despite well in performance, \modelwithspace still faces challenges when inputted with larger digits, highlighting the need for future studies to not only focus on effectiveness and efficiency but also continue to narrow the gap for the LLMs' inability to scale towards larger inputs and the amazing capability in humans.

%% file: data/related.tex
\section{Related Works}

Previous methods that seek to teach LLMs to learn arithmetic mainly focus on the use of step-by-step processes. \textit{Scratchpad}~\cite{nye2021work} was one of the early founders that recognized the use of step-by-step arithmetic solving. \citeauthor{zhou_teaching_2022} focused on in-context learning and showed that a detailed version of \textit{Scratchpad} could significantly improve the accuracy.
\citeauthor{qian2022limitations} recognized the challenger where LLM performance drops as repeated symbols increase.
Goat~\cite{liu2023goat} classified tasks discussed the learnability of different operations and conducted supervised fine-tuning.
\citeauthor{lee2023recursion} proposed the Recursion of Thought to divide the solving process into short contexts.

On the other hand, some works also focus on analyzing arithmetic learning.
\citeauthor{yuan2023large} proposed MATH 401 to evaluate LLM's arithmetic ability. \citeauthor{jelassi2023length} discussed the length generalization ability in arithmetic.
\citeauthor{muffo-etal-2022-evaluating} evaluated the ability of Transformer to perform arithmetic operations following a pipeline that decomposes numbers in decimal before performing computations and demonstrated that this method was 60\% more accurate than GPT-3 on 5-digit addition and subtraction tasks, but was inferior to GPT-3 on 2-digit multiplication tasks. \citeauthor{lee2023teaching} conducted a compressive analysis on training strategies and discussed that reversing the output of addition can speed up the learning process.

%% file: data/conclusion.tex
\section{Conclusion}
In this study, we introduced a novel approach for teaching arithmetic to LLMs by reversing the number order to emphasize the least significant digit. This strategy, which aligns with human arithmetic practices, significantly reduces computational complexity and training data requirements, demonstrating an $11.1\%$ increase in overall accuracy over previous SOTA and showcasing efficiency in token usage during training. The success of our method suggests the potential for broader applications in mathematical problem-solving and in environments with limited resources. We hope this study of ours paves the way for future investigations into optimizing LLM training techniques for numerical reasoning and arithmetic precision.

%% file: data/ethnic.tex
\section*{Limitations}
Our study introduces a novel approach to arithmetic learning in LLMs but is not without limitations. Firstly, our focus on basic arithmetic operations such as addition, subtraction, and multiplication leaves unexplored territories in more complex arithmetic and mathematical problem-solving areas. Secondly, the generalizability of our method to domains beyond arithmetic is yet to be determined. A critical consideration is the reliance on LLMs pretrained with standard numeral expressions; our experiments did not explore the potential benefits of pretraining models directly with reversed numeral expressions. Addressing these limitations could further enhance the applicability and efficiency of LLMs in numerical reasoning and arithmetic precision, suggesting a promising direction for future research to broaden the scope of operations covered and to investigate the impact of pretraining strategies.

\section*{Ethics Statement}
Our research contributes to the field of artificial intelligence by proposing an innovative approach to improve the efficiency and accuracy of LLMs in performing arithmetic operations. This advancement has the potential to positively impact areas where numerical understanding is crucial, including but not limited to, educational technologies, data analysis, and automated reasoning systems. By improving the capability of LLMs to process and understand arithmetic, our work aims to support further developments in technology that can assist in educational settings, enhance scientific research, and provide more reliable computational tools for industries relying on accurate numerical data processing.

We are mindful of the importance of conducting our research with a commitment to ethical principles, ensuring that our methodologies and results are transparent, reproducible, and contribute constructively to the academic community and society at large. While our work primarily focuses on the technical aspects of improving LLMs' arithmetic abilities, we recognize the broader implications of AI and machine learning advancements. Therefore, we encourage the responsible use and continuous ethical evaluation of AI technologies, emphasizing the importance of using such advancements to foster positive societal outcomes.
